\documentclass[letterpaper]{article} 
\usepackage{aaai24}  
\usepackage{times}  
\usepackage{helvet}  
\usepackage{courier}  
\usepackage[hyphens]{url}  
\usepackage{graphicx} 
\usepackage{natbib}  
\usepackage{caption} 
\usepackage{algorithm}
\usepackage{algorithmic}

\usepackage{newfloat}
\usepackage{listings}
\DeclareCaptionStyle{ruled}{labelfont=normalfont,labelsep=colon,strut=off} 
\floatstyle{ruled}
\newfloat{listing}{tb}{lst}{}
\floatname{listing}{Listing}
\title{Block-wise LoRA: Revisiting Fine-grained LoRA for Effective Personalization and Stylization in Text-to-Image Generation }
\author{
    Likun Li\equalcontrib,
    Haoqi Zeng\equalcontrib,
    Changpeng Yang,
    Haozhe Jia\thanks{Corresponding author.},
    Di Xu
}
\affiliations{
    Huawei Cloud Computing Technologies Co., Ltd

    lilikun1@huawei.com,
    zenghaoqi@huawei.com,
    jiahaozhe1@huawei.com
}

\usepackage{bibentry}

\usepackage{amssymb}
\begin{document}

\maketitle

\begin{abstract}
The objective of personalization and stylization in text-to-image is to instruct a pre-trained diffusion model to analyze new concepts introduced by users and incorporate them into expected styles.
Recently, parameter-efficient fine-tuning (PEFT) approaches have been widely adopted to address this task and have greatly propelled the development of this field.
Despite their popularity, existing efficient fine-tuning methods still struggle to achieve effective personalization and stylization in T2I generation. 
To address this issue, we propose block-wise Low-Rank Adaptation (LoRA) to perform fine-grained fine-tuning for different blocks of SD, which can generate images faithful to input prompts and target identity and also with desired style.
Extensive experiments demonstrate the effectiveness of the proposed method.
\end{abstract}

\section{Introduction}




Recently, text-to-image (T2I) generation has been a hot topic in the field of artificial intelligence generation content (AIGC) \cite{ko2023Large-scaleT2I,zhang2023impact}, and lots of diffusion-based generative models have made remarkable progress in this task \cite{ho2020denoising,nichol2021improved,balaji2022ediffi,rombach2022high}.
Among them, Stable Diffusion (SD) \cite{rombach2022latent} can generate realistic and high-quality images corresponding to input prompts, also has been widely used due to its open source nature, and relatively acceptable computing-cost.
However, SD struggles to effectively ingest new character identity or novel style concepts with a low-cost.
Various fine-tuning algorithms such as Dreambooth \cite{ruiz2023dreambooth}, Textual Inversion \cite{gal2022textual_inversion}, and Low-Rank Adaptation (LoRA) \cite{diffusionLoRA} have been proposed to enhance adaptability to specific T2I generation tasks, and have shown convincing capability in personalization and stylization, \textit{e.g.,} creating depictions of Michael Jackson on the moon or transforming the Mona Lisa into comic style.
Among these methods, LoRA stands out and gains extensive adoption among art designers and text-to-image enthusiasts, due to its advantages of low-cost and computational efficiency, making it user-friendly and suitable for consumer devices. 

However, existing LoRA-based methods still fail to address the T2I generations involving personalization and stylization at the same time, which can be observed with inconsistent personal identity and style concept in the generated images.
Moreover, the target concepts and desired style may not be accurately reflected when introducing different LoRAs to an image generation process \cite{gu2023mix,wang2023dynamic}. 
To solve this problem, we propose block-wise LoRA to achieve effective personalization and stylization in T2I generation.
Block-wise LoRA improves original LoRA by skipping some specific blocks of SD during the LoRA-based fine-tuning process for different types of T2I generation tasks. 
Extensive results show that block-wise LoRA can not only reduce the training speed, but also reduce the conflict between different LoRA model, which makes the combination of character ID and style more harmonious.
In addition, we studied the influence of using different U-net blocks to the generation process and gained more understanding of the generation process through the block-wise structure.


\begin{figure*}[t]
\centering
\includegraphics[width=0.9\textwidth]{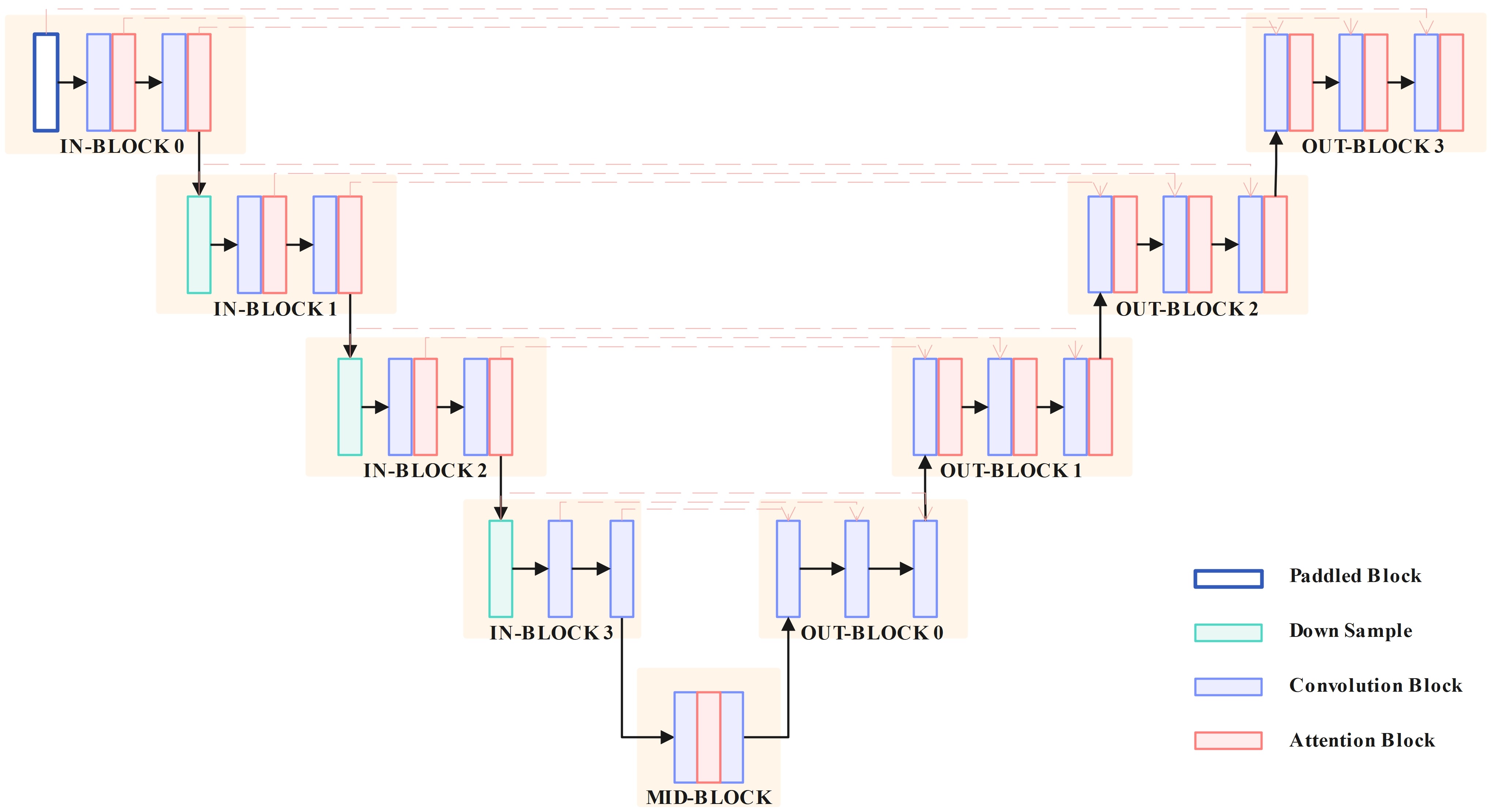} 
\caption{Block partitioning of the U-Net in SD. }
\label{pipeline}
\end{figure*}
\section{Related Work}
\subsection{Parameter-efficient Fine-tuning (PEFT)}
Parameter-efficient fine-tuning (PEFT) focuses on optimizing and adjusting model parameters in a resource-efficient manner. 
This approach aims to enhance the performance of a pre-trained model for specific tasks or domains without requiring an extensive amount of additional data or computational resources. By strategically updating and fine-tuning a minimal set of parameters, this technique achieves efficient adaptation, making it particularly valuable in scenarios with limited labeled data or computational constraints. 
Among many PEFT methods \cite{chen2022adaptformer,gheini2021cross,jie2023fact,karimi2021compacter,lian2022scaling,liu2022few,sung2021training,sung2022lst}, reparameterization-based approaches stand out as the most representative and have been extensively researched \cite{hu2021lora,edalati2022krona,aghajanyan2020intrinsic}.
LoRA \cite{hu2021lora} leverages low-rank approximations to efficiently fine-tune the model parameters, enabling more effective adaptation while mitigating computational demands. 
Specifically tailored for Stable Diffusion models, LoRA offers a promising avenue to enhance adaptability and performance in generative tasks involving diffusion processes.
On this basis, different follow-up approaches \cite{valipour2022dylora,zhang2022adaptive,chavan2023one,aghajanyan2020intrinsic} have been proposed to continuously improve the performance of LoRA.
In this paper, we dig into a fine-grained LoRA by constructing block-wise LoRA adaptors for different blocks of SD and thereby improving the performance of personalization in T2I generation task.

\subsection{T2I Personalization via Fine-tuning}
The goal of T2I personalization is to teach pre-trained models to synthesize novel images with high subject fidelity meanwhile containing a specific style, guided by input prompts.
Many recent works propose different fine-tuning techniques to address this task.
Textual-Inversion-based methods \cite{gal2022image,voynov2023p} inversely translates textual information into personalized visual representations and enhances the adaptability and customization of diffusion-based generative models, offering a means to produce images that closely match the individualized content described in the input text.
DreamBooth and follow-up methods \cite{ruiz2023dreambooth,ruiz2023hyperdreambooth} learn subject-specific prior by optimizing the entire SD network weights, which results in higher subject fidelity in output images. 
Different from this, several works, represented by LoRA \cite{hu2021lora} solve this task through different reparameterizations, which is more light-cost and efficient \cite{kumari2023multi,han2023svdiff,sohn2023styledrop,houlsby2019parameter}.
However, despite generating high-quality subject-driven generations, most of these reparameterization-based methods rarely explore the fine-grained parameters optimization for different blocks of the SD.

\section{Preliminaries}
\noindent\textbf{LoRA} was first proposed as a fine-tuning method for large language models in the field of NLP \cite{hu2021lora} and later introduced to SD models \cite{rombach2022high,podell2023sdxl} for efficient T2I generation. 
LoRA points out that the pre-trained model's weight matrix has a low instrisic feature dimension, and therefore decomposing the model's weight update $\Delta{W}\in\mathbb{R}^{d\times{k}}$ into two low-dimensional matrices, $A\in\mathbb{R}^{r\times k}$ and $B\in\mathbb{R}^{d\times r}$, where $r\ll min(d,k)$ is the dimension of two matrices. 
During the fine-tuning process, the original weights are frozen and only the matrices $A$ and $B$ are tunable. 
As a result, the forward computation process $h=W_{0}x+\Delta Wx$ can be updated as:
\begin{equation}
h=W_{0}x+\Delta Wx=W_{0}x+BAx
\end{equation}
where $h$ is the output feature maps. 
In the early stages of application, LoRA is exclusively used for the cross-attention layers of SD. 
On this basis, LoRA for Convolution Network(LoCon) \cite{LyCORIS} further extends LoRA to convolutional layers. 
For a convolutional layer with weight update $\Delta{W}\in\mathbb{R}^{c_{in}\times{c_{out}}\times{k}\times{k}}$, where $k$ is kernel size, $c_{in}$ and $c_{out}$ are the input and output channel numbers, respectively. 
The convolutional layer can be approximated by two consecutive convolutional layers with weight updates $A\in\mathbb{R}^{r\times{c_{out}}\times{1}\times{1}}$ and $B\in\mathbb{R}^{c_{in}\times{r}\times{k}\times{k}}$ \cite{wang2021pufferfish}.

\begin{figure*}[t]
\centering
\includegraphics[width=1\textwidth]{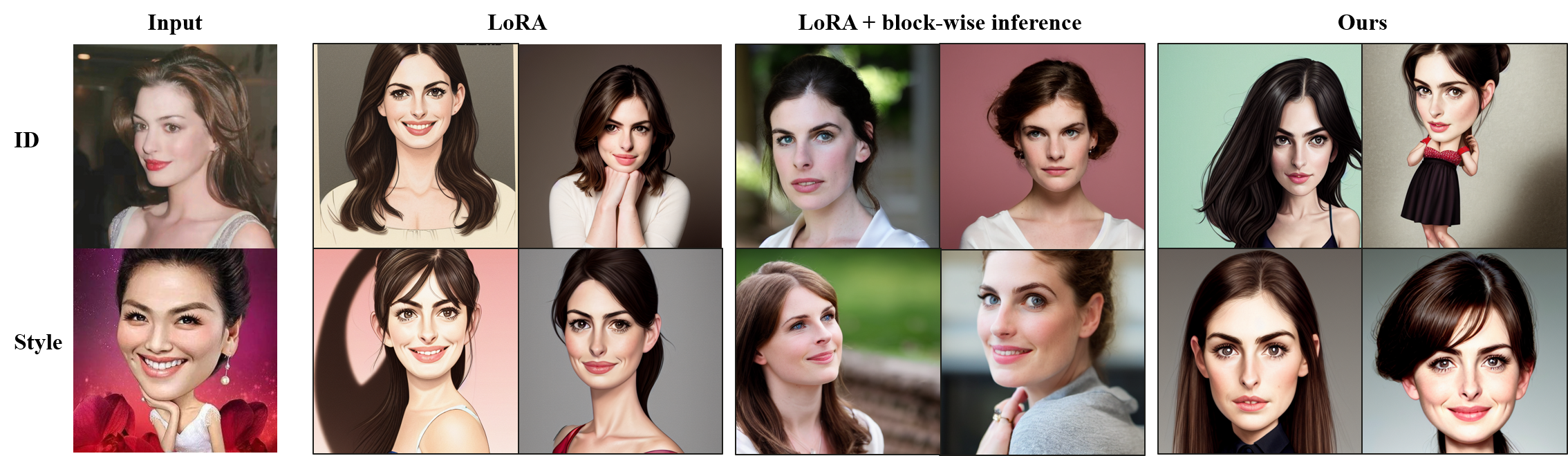}
\caption{Comparisons with LoRA \cite{hu2021lora}. We collect the portraits of Anne Hathaway and cartoon images with a big head style  as the training data for personalization and stylization, respectively. It can be observed that using LoRA fails to generate images with target style, while applying block-wise inference to LoRA struggles to generate corresponding anime images for Hathaway. Our method is capable to achieve expected personalization and stylization in the generated images.
Applying our block-wise training to LoRA can achieve best generation results.}
\label{lora}
\end{figure*}
\begin{figure*}[h]
\centering
\includegraphics[width=1\textwidth]{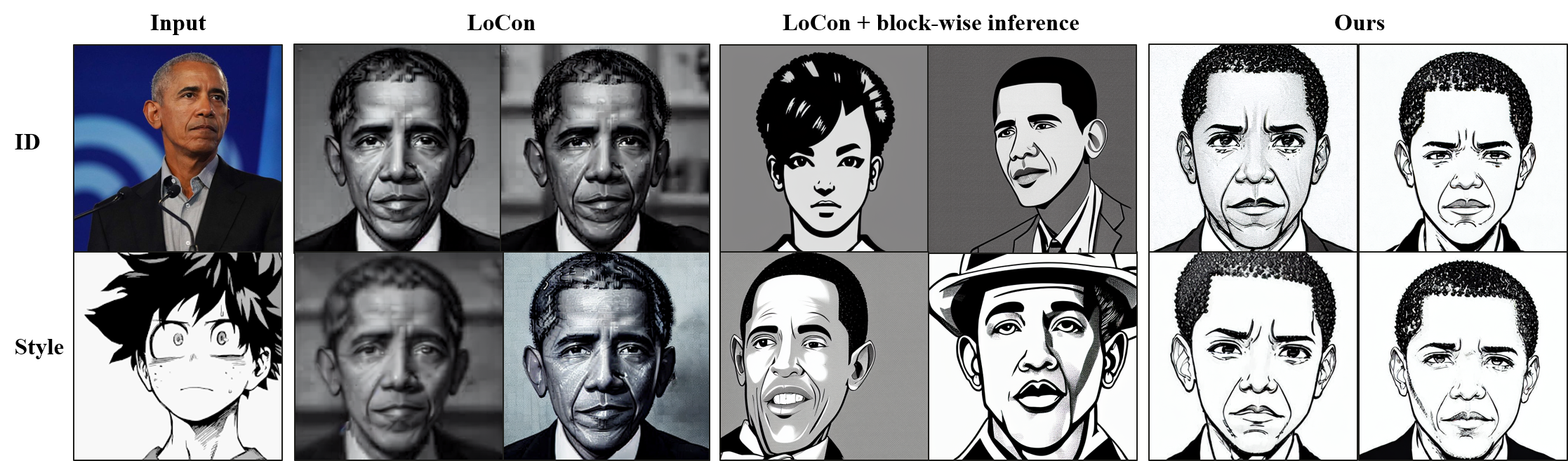}
\caption{Comparisons with LoCon \cite{LyCORIS}. We collect the portraits of Barack Obama and anime images with a specific style  as the training data for personalization and stylization, respectively. Consistent with the comparison results in Figure \ref{lora}, applying our block-wise training to LoCon can still obtain best generation results.}
\label{locon}
\end{figure*}

\section{Method}
\subsection{Block-wise Fine-grained Fine-tuning}
To enhance personalization and stylization in text-to-image generation, we propose a block-wise LoRA to perform fine-grained fine-tuning for SD. 
Generally, LoRA for SD is implemented by performing low-rank fine-tuning attention layers in all blocks of the U-Net in SD (convolutional layers further involved in LoCon). 
However, unsatisfied generation results are often obtained when combining different types of full-blocks-tuned LoRAs for T2I generation.
Therefore, to improve the generation quality of mixed LoRAs, we focus on studying which parts of the U-Net should be fine-tuned for better personalization and stylization.
Specifically, by setting the rank of the LoRA matrix to zero, we can skip the LoRA fine-tuning for current block. 
In this way, SD uses the original pre-trained weights instead of adding LoRA's weight:
\begin{equation}
	\label{eq:alpha}
h =\left\{
           \begin{array}{lcl}
    W_{0}x+BAx & trained \\
    W_{0}x & skipped 
           \end{array}
        \right.
	\end{equation}
Therefore, by controlling LoRA fine-tuning for different parts of U-Net in SD, we can learn the potential impact of block-wise LoRA to the T2I output result.  

As shown in Figure \ref{pipeline}, we divide the U-Net in SD into several parts, including four in-blocks, a mid-block, and four out-blocks.
With this design, we can perform fine-grained low-rank fine-tuning for different blocks of U-Net and evaluate the generation performance of different setting for personalization and sytlization tasks, respectively.
In this work, we opt to train characters using the full-block LoRA/LoCon and conduct style low-rank fine-tuning in block-wise manner, which will be discussed in experiments section.

\section{Experiments}
\begin{figure*}[t]
\centering
\includegraphics[width=1\textwidth]{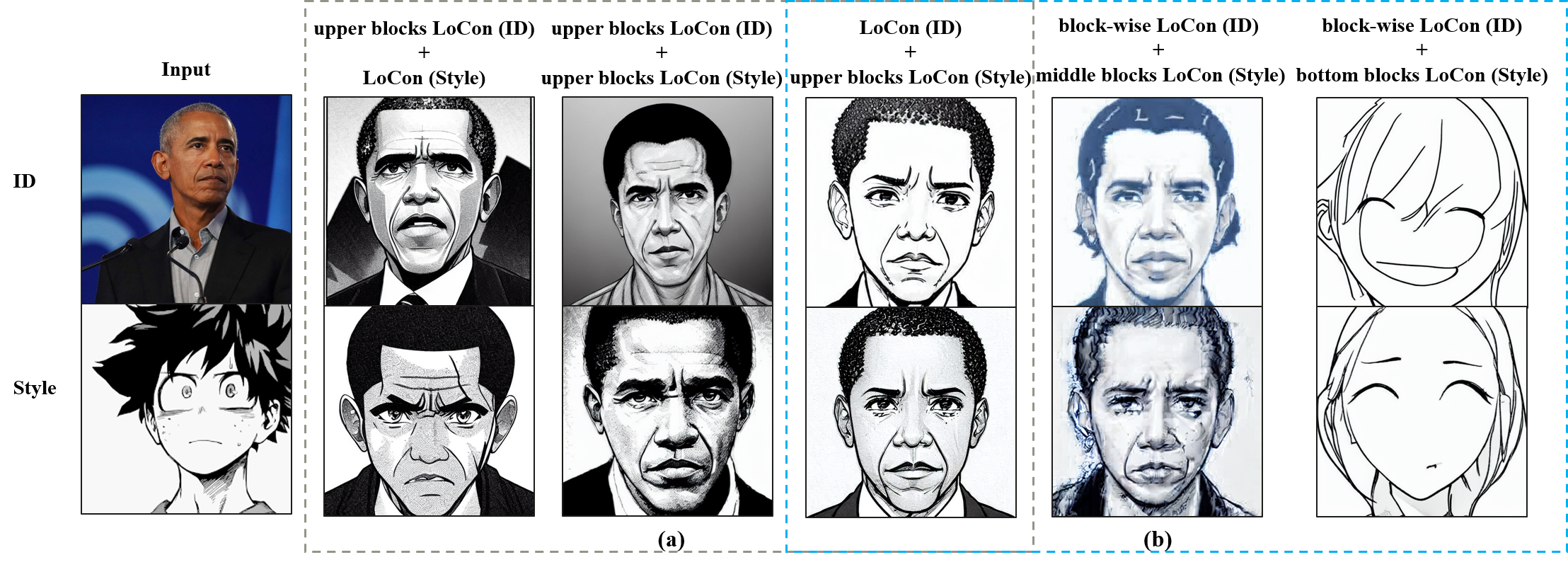}
\caption{Ablation studies of using block-wise LoCon training for personalization and stylization, as well as comparisons between applying block-wise LoCon training for different blocks of SD.}
\label{ablation}
\end{figure*}
\subsection{Implementation details}
We used Manga Face Dataset \cite{mangadataset} as the Manga style LoRA's training dataset. 
For the other style LoRA and character LoRA, we use custom training datasets.
Each dataset is configured to consist of 20 images repeated 25 times, accompanied by approximately 500 regularization images generated by the base model.
According to the work of \cite{LyCORIS}, the training effectiveness of using tags as captions, in the form of $<trigger>, <tag1>, <tag2>, ...<tagN>$, is superior to using natural language captions such as \textit{A photo of S*}. 
Therefore, we employ an annotator \cite{DeepDanbooru} for automated annotation, followed by manual correction of tags to ensure alignment with the content of the images.

For all experiments, we adopted Stable Diffusion 1.4 as the base T2I generation model, and set the fine-tuning steps to 11,000 with a batch size of 2.
During inference, we adopted \textit{DPM 2M++ Karras} as the sampler, with the sampling steps of 25, the classifier-free guidance (CFG) scale of 7.0, and the resolution of generated images consistent with that of the training images.
To conduct a fair comparison, we kept the inference prompts and all hyperparameters fixed for all methods.

\subsection{Results}
First, we compare our proposed block-wise fine-tuning against original LoRA/LoCon , as shown in Figure \ref{lora} and Figure \ref{locon}. 
Here the results of our method were generated with a best setting, which is character LoRA/LoCon + block-wise style LoRA/LoCon, introduced later in the ablation study. 
As the result shows, we can easily see that block-wise LoRA/LoCon models has better personalization and stylization perforamance than the setting of LoRA/LoCon models. 
Meanwhile, the LoRA/LoCon fails to generate images with target style.
In addtion, as the Figure \ref{lora} and Figure \ref{locon} shown, compared to simply using well-tuned LoRA/LoCon for specific blocks of SD during the inference, our proposed block-wise LoRA/LoCon fine-tuning has obvious advantage in personalization. 

In ablation studies,  we first evaluate the performance of three different combinations of different types of character LoCon and Style LoCon.
Each block-wise LoCon combination can mix the character with style. However, as shown in Figure \ref{ablation} (a), block-wise ID LoCon + style LoCon combination's style doesn't match the style LoCon input, block-wise ID LoCon + block-wise style LoCon lost too much character's personal details. Therefore, we consider the ID LoCon + block-wise style LoCon combination has the best result in this set of ablation study, which can not only maintain the character's personal detail, but also change it's style of painting just like the LoCon input.
Moreover,  we explore which blocks should be retained in the combination of ID LoCon + block-wise LoCon. 
First, we separate the block-wise Locon blocks evenly into three parts. 
Then, in every training process, we activate one part of blocks. 
According to Figure \ref{pipeline} shows, the "upper blocks" is In-Block0 + Out-Block3, "middle blocks" is IN-Block1 + Out-Block2 and "bottom blocks" is In-Block2 + Out-Block1.
The results are provided in Figure \ref{ablation} (b).
We can see that when only activate upper blocks, which means the top input blocks and top output blocks, the output image perfectly maintain the detail of character and the style. 
When only activate middle blocks, the output lost it's style, but there are more character's details show up in the picture. 
When only activating bottom blocks, all target information and concepts are missing.

\section{Conclusion}
In this paper, we demonstrate that, when employing multiple LoRA adaptors with SD-based T2I generation model to generate images with target personal identity and style concept, performing LoRA fine-tuning for specified blocks of SD can yield significantly better results than directly performing original LoRA fine-tuning. Moreover, the choice of specific blocks to skip in the LoRA model profoundly influences the T2I model's output. 
Therefore, the block-wise LoRA fine-tuning proposed in this paper can be considered an effective method for generating images when multiple LoRA models collaborate, enhancing personalization and stylization performance. 
Future experiments will focus on combining block-wise LoRA and ControlNet to achieve fine-grained control of visual maps guided T2I generation. 
It is also meaningful to further introduce other reparameterization method such as orthogonal decomposition to block-wise LoRA for a more effective and efficient PEFT.

\bibliography{aaai24}

\begin{thebibliography}{38}
\providecommand{\natexlab}[1]{#1}

\bibitem[{Aghajanyan, Gupta, and Zettlemoyer(2021)}]{aghajanyan2020intrinsic}
Aghajanyan, A.; Gupta, S.; and Zettlemoyer, L. 2021.
\newblock Intrinsic Dimensionality Explains the Effectiveness of Language Model Fine-Tuning.
\newblock In \emph{Proceedings of the 59th Annual Meeting of the Association for Computational Linguistics and the 11th International Joint Conference on Natural Language Processing (Volume 1: Long Papers)}, 7319--7328.

\bibitem[{Balaji et~al.(2022)Balaji, Nah, Huang, Vahdat, Song, Kreis, Aittala, Aila, Laine, Catanzaro et~al.}]{balaji2022ediffi}
Balaji, Y.; Nah, S.; Huang, X.; Vahdat, A.; Song, J.; Kreis, K.; Aittala, M.; Aila, T.; Laine, S.; Catanzaro, B.; et~al. 2022.
\newblock ediffi: Text-to-image diffusion models with an ensemble of expert denoisers.
\newblock \emph{arXiv preprint arXiv:2211.01324}.

\bibitem[{Chavan et~al.(2023)Chavan, Liu, Gupta, Xing, and Shen}]{chavan2023one}
Chavan, A.; Liu, Z.; Gupta, D.; Xing, E.; and Shen, Z. 2023.
\newblock One-for-All: Generalized LoRA for Parameter-Efficient Fine-tuning.
\newblock \emph{arXiv preprint arXiv:2306.07967}.

\bibitem[{Chen et~al.(2022)Chen, Ge, Tong, Wang, Song, Wang, and Luo}]{chen2022adaptformer}
Chen, S.; Ge, C.; Tong, Z.; Wang, J.; Song, Y.; Wang, J.; and Luo, P. 2022.
\newblock Adaptformer: Adapting vision transformers for scalable visual recognition.
\newblock \emph{Advances in Neural Information Processing Systems}, 35: 16664--16678.

\bibitem[{Edalati et~al.(2022)Edalati, Tahaei, Kobyzev, Nia, Clark, and Rezagholizadeh}]{edalati2022krona}
Edalati, A.; Tahaei, M.; Kobyzev, I.; Nia, V.~P.; Clark, J.~J.; and Rezagholizadeh, M. 2022.
\newblock Krona: Parameter efficient tuning with kronecker adapter.
\newblock \emph{arXiv preprint arXiv:2212.10650}.

\bibitem[{Gal et~al.(2022{\natexlab{a}})Gal, Alaluf, Atzmon, Patashnik, Bermano, Chechik, and Cohen-or}]{gal2022textual_inversion}
Gal, R.; Alaluf, Y.; Atzmon, Y.; Patashnik, O.; Bermano, A.~H.; Chechik, G.; and Cohen-or, D. 2022{\natexlab{a}}.
\newblock An Image is Worth One Word: Personalizing Text-to-Image Generation using Textual Inversion.
\newblock In \emph{The Eleventh International Conference on Learning Representations}.

\bibitem[{Gal et~al.(2022{\natexlab{b}})Gal, Alaluf, Atzmon, Patashnik, Bermano, Chechik, and Cohen-or}]{gal2022image}
Gal, R.; Alaluf, Y.; Atzmon, Y.; Patashnik, O.; Bermano, A.~H.; Chechik, G.; and Cohen-or, D. 2022{\natexlab{b}}.
\newblock An Image is Worth One Word: Personalizing Text-to-Image Generation using Textual Inversion.
\newblock In \emph{The Eleventh International Conference on Learning Representations}.

\bibitem[{Gheini, Ren, and May(2021)}]{gheini2021cross}
Gheini, M.; Ren, X.; and May, J. 2021.
\newblock Cross-Attention is All You Need: Adapting Pretrained Transformers for Machine Translation.
\newblock In \emph{Proceedings of the 2021 Conference on Empirical Methods in Natural Language Processing}, 1754--1765.

\bibitem[{Gu et~al.(2023)Gu, Wang, Wu, Shi, Chen, Fan, Xiao, Zhao, Chang, Wu et~al.}]{gu2023mix}
Gu, Y.; Wang, X.; Wu, J.~Z.; Shi, Y.; Chen, Y.; Fan, Z.; Xiao, W.; Zhao, R.; Chang, S.; Wu, W.; et~al. 2023.
\newblock Mix-of-Show: Decentralized Low-Rank Adaptation for Multi-Concept Customization of Diffusion Models.
\newblock \emph{arXiv preprint arXiv:2305.18292}.

\bibitem[{Han et~al.(2023)Han, Li, Zhang, Milanfar, Metaxas, and Yang}]{han2023svdiff}
Han, L.; Li, Y.; Zhang, H.; Milanfar, P.; Metaxas, D.; and Yang, F. 2023.
\newblock Svdiff: Compact parameter space for diffusion fine-tuning.
\newblock \emph{arXiv preprint arXiv:2303.11305}.

\bibitem[{Ho, Jain, and Abbeel(2020)}]{ho2020denoising}
Ho, J.; Jain, A.; and Abbeel, P. 2020.
\newblock Denoising diffusion probabilistic models.
\newblock \emph{Advances in neural information processing systems}, 33: 6840--6851.

\bibitem[{Houlsby et~al.(2019)Houlsby, Giurgiu, Jastrzebski, Morrone, De~Laroussilhe, Gesmundo, Attariyan, and Gelly}]{houlsby2019parameter}
Houlsby, N.; Giurgiu, A.; Jastrzebski, S.; Morrone, B.; De~Laroussilhe, Q.; Gesmundo, A.; Attariyan, M.; and Gelly, S. 2019.
\newblock Parameter-efficient transfer learning for NLP.
\newblock In \emph{International Conference on Machine Learning}, 2790--2799. PMLR.

\bibitem[{Hu et~al.(2021)Hu, Wallis, Allen-Zhu, Li, Wang, Wang, Chen et~al.}]{hu2021lora}
Hu, E.~J.; Wallis, P.; Allen-Zhu, Z.; Li, Y.; Wang, S.; Wang, L.; Chen, W.; et~al. 2021.
\newblock LoRA: Low-Rank Adaptation of Large Language Models.
\newblock In \emph{International Conference on Learning Representations}.

\bibitem[{Jie and Deng(2023)}]{jie2023fact}
Jie, S.; and Deng, Z.-H. 2023.
\newblock Fact: Factor-tuning for lightweight adaptation on vision transformer.
\newblock In \emph{Proceedings of the AAAI Conference on Artificial Intelligence}, volume~37, 1060--1068.

\bibitem[{Karimi~Mahabadi, Henderson, and Ruder(2021)}]{karimi2021compacter}
Karimi~Mahabadi, R.; Henderson, J.; and Ruder, S. 2021.
\newblock Compacter: Efficient low-rank hypercomplex adapter layers.
\newblock \emph{Advances in Neural Information Processing Systems}, 34: 1022--1035.

\bibitem[{Kim(2020)}]{DeepDanbooru}
Kim, K. 2020.
\newblock DeepDanbooru:AI based multi-label girl image classification system, implemented by using TensorFlow.
\newblock \url{https://github.com/KichangKim/DeepDanbooru/tree/master}.

\bibitem[{Ko et~al.(2023)Ko, Park, Jeon, Jo, Kim, and Seo}]{ko2023Large-scaleT2I}
Ko, H.-K.; Park, G.; Jeon, H.; Jo, J.; Kim, J.; and Seo, J. 2023.
\newblock Large-scale text-to-image generation models for visual artists’ creative works.
\newblock In \emph{Proceedings of the 28th International Conference on Intelligent User Interfaces}, 919--933.

\bibitem[{Kumari et~al.(2023)Kumari, Zhang, Zhang, Shechtman, and Zhu}]{kumari2023multi}
Kumari, N.; Zhang, B.; Zhang, R.; Shechtman, E.; and Zhu, J.-Y. 2023.
\newblock Multi-concept customization of text-to-image diffusion.
\newblock In \emph{Proceedings of the IEEE/CVF Conference on Computer Vision and Pattern Recognition}, 1931--1941.

\bibitem[{Köklü(2021)}]{mangadataset}
Köklü, M. 2021.
\newblock Manga Faces Dataset:Manga Faces Dataset classified with facial expressions classes.
\newblock \emph{https://www.kaggle.com/datasets/davidgamalielarcos/manga-faces-dataset/}.

\bibitem[{Lian et~al.(2022)Lian, Zhou, Feng, and Wang}]{lian2022scaling}
Lian, D.; Zhou, D.; Feng, J.; and Wang, X. 2022.
\newblock Scaling \& shifting your features: A new baseline for efficient model tuning.
\newblock \emph{Advances in Neural Information Processing Systems}, 35: 109--123.

\bibitem[{Liu et~al.(2022)Liu, Tam, Muqeeth, Mohta, Huang, Bansal, and Raffel}]{liu2022few}
Liu, H.; Tam, D.; Muqeeth, M.; Mohta, J.; Huang, T.; Bansal, M.; and Raffel, C.~A. 2022.
\newblock Few-shot parameter-efficient fine-tuning is better and cheaper than in-context learning.
\newblock \emph{Advances in Neural Information Processing Systems}, 35: 1950--1965.

\bibitem[{Nichol and Dhariwal(2021)}]{nichol2021improved}
Nichol, A.~Q.; and Dhariwal, P. 2021.
\newblock Improved denoising diffusion probabilistic models.
\newblock In \emph{International Conference on Machine Learning}, 8162--8171. PMLR.

\bibitem[{Podell et~al.(2023)Podell, English, Lacey, Blattmann, Dockhorn, M{\"u}ller, Penna, and Rombach}]{podell2023sdxl}
Podell, D.; English, Z.; Lacey, K.; Blattmann, A.; Dockhorn, T.; M{\"u}ller, J.; Penna, J.; and Rombach, R. 2023.
\newblock Sdxl: Improving latent diffusion models for high-resolution image synthesis.
\newblock \emph{arXiv preprint arXiv:2307.01952}.

\bibitem[{Rombach et~al.(2022{\natexlab{a}})Rombach, Blattmann, Lorenz, Esser, and Ommer}]{rombach2022high}
Rombach, R.; Blattmann, A.; Lorenz, D.; Esser, P.; and Ommer, B. 2022{\natexlab{a}}.
\newblock High-resolution image synthesis with latent diffusion models.
\newblock In \emph{Proceedings of the IEEE/CVF conference on computer vision and pattern recognition}, 10684--10695.

\bibitem[{Rombach et~al.(2022{\natexlab{b}})Rombach, Blattmann, Lorenz, Esser, and Ommer}]{rombach2022latent}
Rombach, R.; Blattmann, A.; Lorenz, D.; Esser, P.; and Ommer, B. 2022{\natexlab{b}}.
\newblock High-resolution image synthesis with latent diffusion models.
\newblock In \emph{Proceedings of the IEEE/CVF conference on computer vision and pattern recognition}, 10684--10695.

\bibitem[{Ruiz et~al.(2023{\natexlab{a}})Ruiz, Li, Jampani, Pritch, Rubinstein, and Aberman}]{ruiz2023dreambooth}
Ruiz, N.; Li, Y.; Jampani, V.; Pritch, Y.; Rubinstein, M.; and Aberman, K. 2023{\natexlab{a}}.
\newblock Dreambooth: Fine tuning text-to-image diffusion models for subject-driven generation.
\newblock In \emph{Proceedings of the IEEE/CVF Conference on Computer Vision and Pattern Recognition}, 22500--22510.

\bibitem[{Ruiz et~al.(2023{\natexlab{b}})Ruiz, Li, Jampani, Wei, Hou, Pritch, Wadhwa, Rubinstein, and Aberman}]{ruiz2023hyperdreambooth}
Ruiz, N.; Li, Y.; Jampani, V.; Wei, W.; Hou, T.; Pritch, Y.; Wadhwa, N.; Rubinstein, M.; and Aberman, K. 2023{\natexlab{b}}.
\newblock Hyperdreambooth: Hypernetworks for fast personalization of text-to-image models.
\newblock \emph{arXiv preprint arXiv:2307.06949}.

\bibitem[{Ryu(2022)}]{diffusionLoRA}
Ryu, S. 2022.
\newblock Low-rank adaptation for fast text-to-image diffusion fine-tuning.
\newblock \url{https://github.com/cloneofsimo/lora}.

\bibitem[{Sohn et~al.(2023)Sohn, Ruiz, Lee, Chin, Blok, Chang, Barber, Jiang, Entis, Li et~al.}]{sohn2023styledrop}
Sohn, K.; Ruiz, N.; Lee, K.; Chin, D.~C.; Blok, I.; Chang, H.; Barber, J.; Jiang, L.; Entis, G.; Li, Y.; et~al. 2023.
\newblock StyleDrop: Text-to-Image Generation in Any Style.
\newblock \emph{arXiv preprint arXiv:2306.00983}.

\bibitem[{Sung, Cho, and Bansal(2022)}]{sung2022lst}
Sung, Y.-L.; Cho, J.; and Bansal, M. 2022.
\newblock Lst: Ladder side-tuning for parameter and memory efficient transfer learning.
\newblock \emph{Advances in Neural Information Processing Systems}, 35: 12991--13005.

\bibitem[{Sung, Nair, and Raffel(2021)}]{sung2021training}
Sung, Y.-L.; Nair, V.; and Raffel, C.~A. 2021.
\newblock Training neural networks with fixed sparse masks.
\newblock \emph{Advances in Neural Information Processing Systems}, 34: 24193--24205.

\bibitem[{Valipour et~al.(2023)Valipour, Rezagholizadeh, Kobyzev, and Ghodsi}]{valipour2022dylora}
Valipour, M.; Rezagholizadeh, M.; Kobyzev, I.; and Ghodsi, A. 2023.
\newblock DyLoRA: Parameter-Efficient Tuning of Pre-trained Models using Dynamic Search-Free Low-Rank Adaptation.
\newblock In \emph{Proceedings of the 17th Conference of the European Chapter of the Association for Computational Linguistics}, 3266--3279.

\bibitem[{Voynov et~al.(2023)Voynov, Chu, Cohen-Or, and Aberman}]{voynov2023p}
Voynov, A.; Chu, Q.; Cohen-Or, D.; and Aberman, K. 2023.
\newblock $ P+ $: Extended Textual Conditioning in Text-to-Image Generation.
\newblock \emph{arXiv preprint arXiv:2303.09522}.

\bibitem[{Wang, Agarwal, and Papailiopoulos(2021)}]{wang2021pufferfish}
Wang, H.; Agarwal, S.; and Papailiopoulos, D. 2021.
\newblock Pufferfish: Communication-efficient models at no extra cost.
\newblock \emph{Proceedings of Machine Learning and Systems}, 3: 365--386.

\bibitem[{Wang et~al.(2023)Wang, Yang, Yang, Butt, and van~de Weijer}]{wang2023dynamic}
Wang, K.; Yang, F.; Yang, S.; Butt, M.~A.; and van~de Weijer, J. 2023.
\newblock Dynamic Prompt Learning: Addressing Cross-Attention Leakage for Text-Based Image Editing.
\newblock \emph{arXiv preprint arXiv:2309.15664}.

\bibitem[{Yeh et~al.(2023)Yeh, Hsieh, Gao, Yang, Oh, and Gong}]{LyCORIS}
Yeh, S.-Y.; Hsieh, Y.-G.; Gao, Z.; Yang, B. B.~W.; Oh, G.; and Gong, Y. 2023.
\newblock Navigating Text-To-Image Customization: From LyCORIS Fine-Tuning to Model Evaluation.
\newblock arXiv:2309.14859.

\bibitem[{Zhang, Kwon, and Xiong(2023)}]{zhang2023impact}
Zhang, K.; Kwon, O.; and Xiong, H. 2023.
\newblock The Impact of Generative Artificial Intelligence.
\newblock Technical report, arXiv. org.

\bibitem[{Zhang et~al.(2022)Zhang, Chen, Bukharin, He, Cheng, Chen, and Zhao}]{zhang2022adaptive}
Zhang, Q.; Chen, M.; Bukharin, A.; He, P.; Cheng, Y.; Chen, W.; and Zhao, T. 2022.
\newblock Adaptive Budget Allocation for Parameter-Efficient Fine-Tuning.
\newblock In \emph{The Eleventh International Conference on Learning Representations}.

\end{thebibliography}

\end{document}